\documentclass{article}

\usepackage{arxiv}

\usepackage[utf8]{inputenc} 
\usepackage[T1]{fontenc}    
\usepackage{hyperref}       
\usepackage{url}            
\usepackage{booktabs}       
\usepackage{amsfonts}       
\usepackage{nicefrac}       
\usepackage{microtype}      
\usepackage{lipsum}

\usepackage[colorinlistoftodos]{todonotes}
\usepackage{algorithm}
\usepackage{algpseudocode}
\usepackage{multirow}

\usepackage{color}
\usepackage{graphicx}
\usepackage{subcaption}
\usepackage{enumitem}
\usepackage{multirow}
\usepackage{mathtools,mathptmx}
\usepackage{xcolor}

\newcommand{\MM}{Minimax}
\newcommand{\bigO}{\mathcal{O}}
\newcommand{\cmmnt}[1]{\ignorespaces}

\title{Memory-Efficient Sampling for Minimax Distance Measures}

\author{
  Fazeleh Sadat Hoseini\\
  Department of Computer Engineering\\
  Chalmers University of Technology\\
  Göteborg. Sweden \\
  \texttt{fazeleh@chalmers.se} \\
   \And
  Morteza Haghir Chehreghani \\
  Department of Computer Engineering\\
  Chalmers University of Technology\\
  Göteborg, Sweden \\
  \texttt{morteza.chehreghani@chalmers.se} \\
}

\begin{document}
\maketitle

\begin{abstract}
Minimax distance measure extracts the underlying patterns and manifolds in an unsupervised manner. The existing methods require a quadratic memory with respect to the number of objects.

In this paper, we investigate efficient sampling schemes in order to reduce the memory requirement and provide a linear space complexity. In particular, we propose a novel sampling technique that adapts well with Minimax distances. We evaluate the methods on real-world datasets from different domains and analyze the results.
\end{abstract}



\section{Introduction}


Learning a proper representation is usually the first step in every machine learning and data analytic tasks.
Some recent representation learning methods have been developed in the context of deep learning \cite{DeepLearnig}, which are highly parameterized and require  a huge amount of labeled data for training. On the other hand, there are methods that learn a proper representation in an unsupervised way and usually do not require learning free parameters. 


A category of unsupervised representations and distance measures, called \emph{link-based} distance \cite{Fouss:2012,Chebotarev:2011}, take into account all the \emph{paths} between the objects represented in a graph. These distance measures are often obtained by inverting the Laplacian of the base distance matrix in the context of Markov diffusion kernel \cite{Fouss:2012}.
However, computing link-based distances for all the pairs of objects requires $ \bigO \left ( N^3 \right )$ runtime, where $N$ is the number of objects. This complexity might not be affordable for very large datasets.
%
%
A more sophisticated distance measure, called \emph{\MM\ distance} (or \emph{path-based} distance \cite{FischerB03}), seeks for the minimum largest gap among all feasible paths between the objects represented in a graph \cite{chehreghani2019nonparametric}. 
\MM\ distance measure, i) detects the underlying characteristics of the data in an unsupervised way, ii) extracts elongated structures and manifolds via considering transitive relations,  and iii)  appropriately adapts to the shape of different structures. 
It has been used in several clustering, classification and matrix (of user profiles) completion applications \cite{FischerB03,KimC13,ChehreghaniSDM16,Chehreghani17ECIR, chehreghani2019nonparametric}.

Computing pairwise \MM\ distances can be expensive with respect to both computations (runtime) and memory.
The first pairwise \MM\ distance methods were based on some variants of the Floyd-Warshall algorithm with the computational complexity of $ \bigO \left ( N^3 \right )$ \cite{Cormen:2001:IA:580470}. Recently more efficient methods have been developed for computing pairwise \MM\ distances \cite{chehreghani2019nonparametric,ChehreghaniAAAI17}, even for sparse data \cite{ChehreghaniICDM17}. 
$K$-nearest neighbor search is another task wherein the \MM\ distances have been applied successfully and several computationally efficient methods have been developed for that \cite{KimC13,ChehreghaniSDM16}.

Despite the aforementioned recent advances in computational efficiency of \MM\ distances, the memory efficiency and space complexity have not been properly studied yet. 
Computing pairwise \MM\ distances requires $ \bigO \left ( N^2 \right )$ memory. However, such a memory capacity might be unavailable in many applications. For example, embedded devices and systems can afford only a limited amount of memory. Therefore, in this paper, we study a memory-efficient solution for pairwise \MM\ distance measures.



We develop a generic framework for efficient sampling schemes that require only $ \bigO \left ( N \right )$ memory (space) complexity for computing pairwise  \MM\ distances, instead of $ \bigO \left ( N^2 \right )$. Within this framework, in particular, we propose a sampling method that is adaptive to \MM\ distances. This method computes the samples that are maximally consistent with the  \MM\ distances on the original dataset. We also investigate other sampling methods based on $k$-means, Determinantal Point Process (DPP) and random sampling. We evaluate the framework on several real-world and synthetic datasets with a clustering such as Gaussian Mixture Model (GMM). 

\section{Background and Definitions}
A dataset is described by a set of objects with the indices $\mathbf{O} = \{1,..., N\}$, and the respective measurements $\mathbf{X}$. The measurements can be the feature vectors of different objects.
We assume a function $f\left(i,j \right)$ that gives the pairwise dissimilarity between objects $i$ and $j$. Thus, the dataset can be shown by a graph $ G\left( \mathbf{O},\mathbf{W}\right)$, where the nodes $\mathbf{O}$ represent the objects and the edge weights $\mathbf{W}$ show the pairwise dissimilarities obtained by $f$. Note that in some applications, the graph  $G\left(\mathbf{O}, \mathbf{W} \right)$ might be directly given without the explicit measurements $\mathbf{X}$. $\mathbf{W}$ does not need to satisfy all properties of a metric, i.e., the triangle inequality might be violated. The assumptions for $\mathbf{W}$ are: i) zero self-dissimilarities, i.e., $f\left(i,i\right) =0$, ii) symmetry, i.e., $f\left(i,j \right) =f\left(j,i \right)$, and iii) non-negativity, i.e., $\forall i,j: f\left(j,i\right)\ge 0$.

The \MM\ distance between $i,\ j \in  \mathbf O$, denoted by $\mathbf M_{ij}$ is defined as
\cite{FischerB03,chehreghani2019nonparametric}

\begin{equation}\label{eq:MM}
    \mathbf M_{ij} = \min_{p \in P_{i,j}} \max_{e\in p} \mathbf W_{e_1,e_2}
\end{equation}

where $P_{i,j}$ is the set of paths connecting $i$ and $j$ over $G\left(\mathbf O, \mathbf W\right)$, and $e_1$ and $e_2$ are two ends of edge $e$ in path $p$. 
A path $p$ is characterized the set of the consecutive edges on that. 

As shown in Eq. \ref{eq:MM}, to compute the \MM\ distance between two objects, we need to trace all possible paths between them. A modified variant of the Floyd–Warshall algorithm with a computational complexity of $\bigO\left(N^3\right)$ can be used to find the pairwise \MM\ distances. 
As studied in \cite{ChehreghaniAAAI17,chehreghani2019nonparametric}, given graph $G\left(\mathbf O, \mathbf W \right)$, 
a minimum spanning tree (MST) over $\mathbf{G}$ provides the necessary and sufficient information to obtain the pairwise \MM\  distances. Thus, the pairwise \MM\ distances on an arbitrary graph are equal to the pairwise \MM\ distances computed on any MST over the graph.  This result can reduce the computational complexity of computing pairwise \MM\  distances to $\bigO\left(N^2\right)$ \cite{ChehreghaniAAAI17}.

\section{Sampling for Minimax Distances}
\subsection{Sampling Framework}
We propose a memory-efficient sampling framework for computing pairwise \MM\ distances.
We assume that the available memory is linear in the number of objects, i.e., the space complexity should be $\bigO \left(N\right)$. This makes our approach applicable for low-memory settings, like embedded devices with limited memory capacity.


\begin{algorithm}[t]
	\caption{Memory-efficient computation of \MM\ distances}
	\label{alg:overview}
	\begin{algorithmic}[1]
	    \State Compute $\sqrt{N}$ samples $\mathbf S$ from the entire dataset.
	    \State Obtain the pairwise \MM\ distances $\mathbf M_s$ of the objects in $\mathbf S$.
	    \State Compute an embedding of $\mathbf M_s$ into an Euclidean space.
	    \State Apply the clustering algorithm on $\sqrt{N}$ embedded samples.
	    \State Extend the samples' labels in $\mathbf S$ to all the object in $\mathbf O$.
	\end{algorithmic} 
\end{algorithm}

Algorithm \ref{alg:overview} describes the different steps of the framework. 
We first compute $\sqrt{N}$ samples from the data and store the sample indices in $\mathbf S$. Then, we compute the pairwise \MM\ distances of the samples in $\mathbf M_s$. The memory needed for these pairwise distances would be $\bigO \left(\sqrt N \times \sqrt N\right) = \bigO \left(N \right)$. 

While clustering methods such as spectral clustering and kernel $k$-means can be applied to the pairwise relations, methods like Gaussian Mixture Models (GMM) require vector (feature) representation of the measurements. Vectors constitute the most basic data representation that many numerical machine learning methods can be applied to them. Thus, we employ an embedding of the pairwise  \MM\ distances of the samples in $\mathbf M_s$ into an Euclidean space such that the squared Euclidean distances in the new space equal the pairwise \MM\ distances in $\mathbf M_s$.
The feasibility of such an embedding has been shown in \cite{ChehreghaniAAAI17} by studying the \emph{ultrametric} properties of  pairwise \MM\ distances. Hence we can employ an Euclidean embedding as following \cite{RePEc1938}. First we transform $\mathbf M_s$ into a Mercer kernel, which is positive semidefinite (PSD) and symmetric. So we can decompose it to its eigenvectors $\mathbf V = (\mathbf v_1,...,\mathbf v_{\sqrt N})$ and the diagonal eigenvalue matrix $\Lambda = \texttt{diag}\big(\lambda_{1},..., \lambda_{\sqrt N}\big)$ where the eigenvalues are sorted in decreasing order. For a given dimensionality $d^{'} \in \left \{ 1, ... , \sqrt N \right \}$, the embedded $\sqrt N \times d^{'}$ matrix, denoted by $\mathbf E_{d^{'}}$, is computed by
\begin{equation}\label{eq:embedding}
    \mathbf E_{d^{'}} = \Big( \mathbf V_{\big[1..d^{'}\big] }
                \Big( \Lambda_{\big[1..d^{'}\big] } \Big)  ^{\frac{1}{2}}
\end{equation}

It is notable that the space complexity of this embedding is $\bigO \left(\sqrt N \times \sqrt N\right) = \bigO \left(N\right)$, since $d^{'} \le \sqrt N$.

With sampling, we divide the dataset into $|\mathbf S|$ disjoint subsets. In other words, each sample represents a group of objects,  and every object is represented by one and only one sample. Owing to this, after applying a clustering method on the samples $\mathbf S$, we extend each sample's label to the objects represented by that particular sample. Finally, we evaluate the results over the entire dataset.


\subsection{\MM \ Sampling}

\begin{algorithm}[t]
   \caption{ \MM \ (MM) Sampling}\label{alg:MM}
	\begin{algorithmic}[1]
	    \State $\mathbf T \gets $ Perform incremental  Prim's MST algorithm. 
	    \State Permute rows of $\mathbf T$ such that $\mathbf{T}_{:,1}$ appears in ascending order.
	    \State  Initialize $SubsetID_j$ by $j, \forall j \in \mathbf O$. 
	    \For{$i = 1\ to\ N-\sqrt{N} $}  \Comment{\textcolor{blue}{such that at the end only $\sqrt{N}$ subsets are left.}}
	        \State $e_1\gets \mathbf{T}_{i,2}$ and $e_2\gets\mathbf{T}_{i,3}$.
	        \State $ID_1\gets SubsetID_{e_1}$ and $ID_2\gets SubsetID_{e_2}$.
	    	\State $ new\_ID \ \gets \ $ Generate a new ID.
	    	\State $SubsetID_j =\ new\_ID \ , \forall j \in \mathbf O$ such that
	    	$SubsetID_j =  ID_1$ \\ $\qquad\qquad\qquad\qquad\qquad\qquad\qquad \qquad$ \ or \  $SubsetID_j = ID_2$.
	    	
		\EndFor
		\State  \textbf{Return} the final $\sqrt{N}$ subsets as the samples. 
	\end{algorithmic} 
\end{algorithm}

Within our sampling framework, we in particular propose an effective memory-efficient sampling method well-adapted to the \MM\ distances. We call this method \MM\ Sampling or MM Sampling in short.  Algorithm \ref{alg:MM} describes its  different steps.

We first compute a minimum spanning tree (MST) on the given dataset. Such a minimum spanning tree, as studied in \cite{ChehreghaniAAAI17,chehreghani2019nonparametric}  contains the sufficient information for computing \MM\ distances. Thus we can establish our sampling scheme based on that. Different MST algorithms, usually assume a graph of data like $G\left(\mathbf O, \mathbf W\right)$ is given, and then compute the minimum spanning tree over that. However, such a graph can require an $\bigO \left(N^2 \right)$ space complexity, which cannot be afforded in our setting. 


Therefore, to compute the minimum spanning tree, we employ the Prim's algorithm in an incremental way. Let list $\mathbf C$ shows the current objects in the MST, that is initialized by a random object $v$, i.e., $\mathbf C = \big\{v\big\}$. Also, vector $\mathbf{l}$ of size $N$ indicates the dissimilarities between $\mathbf C$ and each of the objects in $\mathbf{O} \setminus \mathbf C$. At the beginning, $\mathbf{l}$ is initialized with the pairwise dissimilarities between $v$ and the objects in $\mathbf{O}$ (obtained by $f$). We set the  elements in $\mathbf{l}$ to $\text{+}\infty$ if the respective indices are in $\mathbf C$, i.e., if $i \in \mathbf C$, then $\mathbf{l}_i = \text{+}\infty$. Thus at the beginning we have $\mathbf{l}_v = \text{+}\infty$. 
Then, Prim's algorithm at each step, i) obtains the object index with a  minimal element in $\mathbf{l}$ (i.e., $u := \arg\min_i \mathbf{l}_i$), ii) adds $u$ to $\mathbf C$ (i.e., $\mathbf C = \mathbf C \cup \big\{u\big\}$), iii) sets $\mathbf{l}_u = \text{+}\infty$, and iv) updates the dissimilarities between the unselected objects and  $\mathbf C$, i.e., $\text{if } \mathbf{l}_i \neq \text{+}\infty, \text{ then } \mathbf{l}_i = \min\big(\mathbf{l}_i, f\big(u,i\big)\big)$. These steps are repeated $N-1$ until a complete MST is built. We notice that all the used data structures are at most linear in size, i.e., the space complexity of this stage is $\bigO \left(N\right)$. The output will be the $\big(N-1\big) \times 3$ matrix $\mathbf T$, wherein the first column (i.e., $\mathbf T_{:,1}$) indicates the weights (dissimilarities) of the edges added one by one to the MST, and the second and third columns ($\mathbf T_{:,2}$ and $\mathbf T_{:,3}$) store the two end points at the two sides of the respective MST edges.

We then permute the rows of $\mathbf T$ such that the edge weights in the first column $\mathbf T_{:,1}$ appear in ascending order. This can be done via an in-place sorting method without a need for significant extra memory \cite{Cormen:2001:IA:580470}. 

In the next stage, we construct subsets (components) of objects. Then, each final subset will serve as a sample. 
We start with each object in a separate subset with its own ID in \emph{SubsetID}. \emph{SubsetID} is a vector of size $N$, wherein $SubsetID_i$ shows the ID of the latest subset that the object $i$ belongs to.  $SubsetID_i$ is initialized by $i$. 
Thus, the algorithm starts with the number of subsets equal to number of objects $N$. The algorithm performs iteratively on the sorted edge weights $\mathbf T_{:,1}$ in the MST. In iteration $\mathit{i}$, we consider the edge weight $\mathbf{T}_{i,1}$ and combine the two subsets that the two objects at the two sides of the edge belong to. Let $\mathbf{s1}$ shows the objects that have the same \emph{SubsetID} as the object in $\mathbf T_{i,2}$ and  $\mathbf{s2}$ shows the objects that have the same \emph{SubsetID} as the object in $\mathbf T_{i,3}$. 
By combining $\mathbf{s1}$ and $\mathbf{s2}$ we create a new subset. 
We consider a new ID to this newly created subset (by increasing a global variable that enumerates the subsets) and  we update the \emph{SubsetID} of all the objects in the new subset with this new ID.
The algorithm iterates until the number of (nonempty) subsets becomes equal to the desired number of samples, i.e., $\sqrt{N}$.


As mentioned, a MST sufficiently contains the edges that represent the pairwise \MM\ distances. Thus, when we combine two subsets, then the weight  of the respective edge represents the \MM\ distance between the set of objects in $\mathbf{s1}$ and the set of objects in $\mathbf{s2}$. Because, i) this edge weight is the largest dissimilarity on the (only) MST path between the objects in  $\mathbf{s1}$ and those in $\mathbf{s2}$, ii) all the other edge weights inside the subsets are smaller (or equal) as they are visited earlier. This implies that our sampling method computes the subsets (samples) such that internal \MM\ distances (intra-sample \MM\ distances) are kept minimal. In other words, we discard the largest $\sqrt{N}-1$ \MM\ distances to produce $\sqrt{N}$ samples. Thus, this method is adaptive and consistent with the \MM\ distances on entire data.

Next, we need to compute the pairwise \MM\ distances between the $\sqrt{N}$ subsets (samples) to obtain $\mathbf M_s$.
According to the aforementioned argument, after computing the final $\sqrt{N}$ subsets, we can continue the procedure in Algorithm \ref{alg:MM}, but for the purpose of computing the pairwise \MM\ distances between the subsets. By performing the procedure until only one subset is left, we obtain  $\mathbf M_s$, where at each iteration, we set the \MM\ distances between the final subsets in $\mathbf{s1}$ and $\mathbf{s2}$ by the visited edge weight (after a proper reindexing).
We note that the steps of Algorithm \ref{alg:MM} require at most linear memory w.r.t. $N$. 



\subsection{Other Sampling Methods}
Our generic sampling framework in Algorithm \ref{alg:overview} allows us to investigate different sampling strategies for  pairwise \MM\ distances. Here, we study for example two other sampling methods based on $k$-means and Detrimental Point Process (DPP). 
Random sampling is another choice that selects $\sqrt{N}$ objects uniformly at random.
\begin{itemize}[leftmargin=*]
    \item \textbf{$k$-means sampling:} This method is grounded on $k$-means clustering with space complexity of $\bigO \Big( N \big( D$\textit{+}$ K \big) \Big) $, where $N$ is number of objects, $D$ is data dimensionality, and $k$ is the number of clusters . To obtain $\sqrt{N}$ samples, we apply $k$-means clustering with $k$ equal to $\sqrt{N}$. Then, we consider the centroids of the clusters as the samples $\mathbf S$, where each sample represents the objects belonging to the cluster.  
    \item \textbf{Sampling with Determinantal Point Processes (DPP):} DPP \cite{DPP} is a  probabilistic model that relies on a similarity measure between the objects. This model advocates \emph{diversity} in samples in addition to \emph{relevance}. A point process $ P $ on a ground discrete set $ \mathcal{Y} =\left\{ 1,\ 2,\ ...,\ N\right\}$  is a probability measure over point configuration of all subset of $\mathcal{Y}$, denoted by $2^{\mathcal{Y}}$.  $P $ is a determinantal point process if $\mathbf{Y}$ is a random subset drawn according to $P$, then  for every $A \subseteq  \mathcal{Y}$ we have $\ P \left(A \subseteq \mathcal{Y} \right) = det\left(\mathbf K_A \right)$. $\mathbf K$ is a positive semidefinite matrix, called marginal kernel, where $\mathbf K_{ij}$  indicates the similarity of objects $i$ and $j$ \cite{DPP}. In the standard DPP, the space complexity of $\mathbf K$ is $\bigO \big(N^2\big)$ which is a drawback of the DPP sampling. Thus, one may assume these samples are computed offline and then are provided to the memory-constrained system.   
    
\end{itemize}

For both cases, computing the \MM\ distances among the samples is memory-efficient, as we have only $\sqrt N$ samples.



\section{Experimental Results}

In this section, we demonstrate the performance of our framework on clustering of several synthetic and real-world datasets.
We apply the different sampling methods on three synthetic datasets \cite{ClusteringDatasets} (Pathbased, Spiral, Aggregation), five real-world datasets from UCI repository \cite{UCI:2019} (Banknote Authentication, Cloud, Iris, Perfume, Seed), and two images  from PASCAL VOC collection \cite{pascal-voc-2012} with the indices 2009\_002204 and 2011\_001967. Whenever needed, we compute the pairwise dissimilarities (i.e., $f\left(i,j \right)$) according to the squared Euclidean distance between the respective vectors.

For each dataset, we follow the proposed steps to compute a sample set and then extend the solution to the objects in the entire dataset. For clustering, we use GMM, while additional experiments with $k$-means yield consistent results.
For these datasets, we have access to the ground truth. Thus to evaluate the results, we use these three commonly used  metrics: Rand score, mutual information, and v-measure denoted by M1, M2, and M3 which indicate the similarity, agreement, and homogeneity of estimated labels and true labels.

\begin{figure}[H]
\begin{center}
\begin{minipage}[b]{.55\linewidth}
    \begin{subfigure}[b]{0.49\textwidth}
    \centering
    \captionsetup{justification=centering}
        \includegraphics[width=\textwidth]{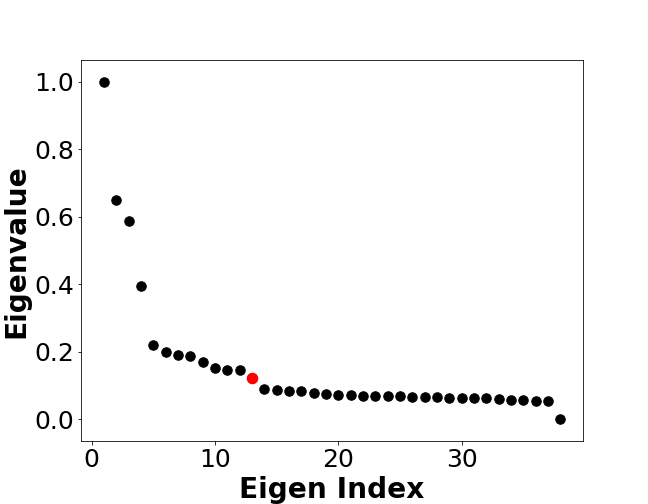}
        \caption{\footnotesize MM Sampling}
        \label{fig:heatmapXeonPhiReal}
    \end{subfigure}
    \begin{subfigure}[b]{0.49\textwidth}
    \centering
    \captionsetup{justification=centering}
        \includegraphics[width=\textwidth]{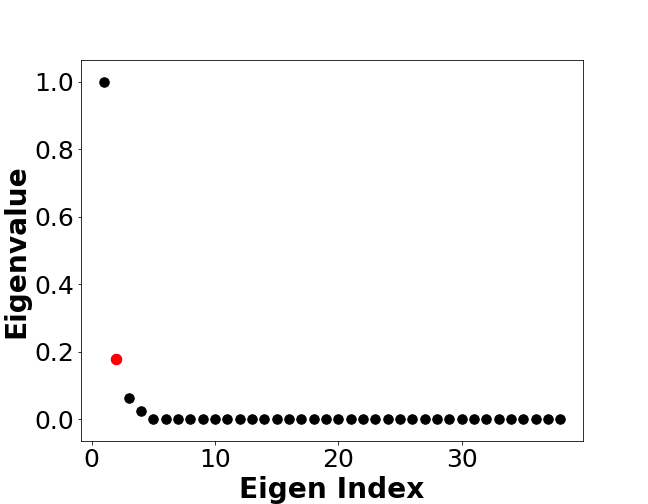}
        \caption{ \footnotesize   k-means Sampling}
        \label{fig:heatmapXeonPhiModel}
    \end{subfigure}
    \newline
    \begin{subfigure}[b]{0.49\textwidth}
    \centering
    \captionsetup{justification=centering}
        \includegraphics[width=\textwidth]{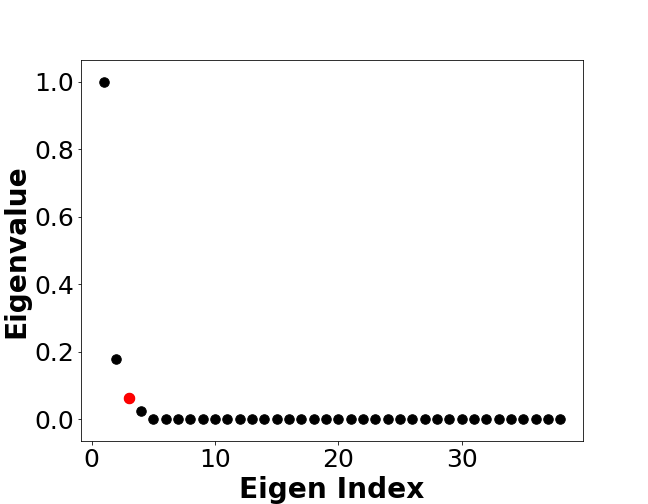}
        \caption{\footnotesize  DPP Sampling}
        \label{fig:heatmapXeonPhiReal}
    \end{subfigure}
    \begin{subfigure}[b]{0.49\textwidth}
    \centering
    \captionsetup{justification=centering}
        \includegraphics[width=\textwidth]{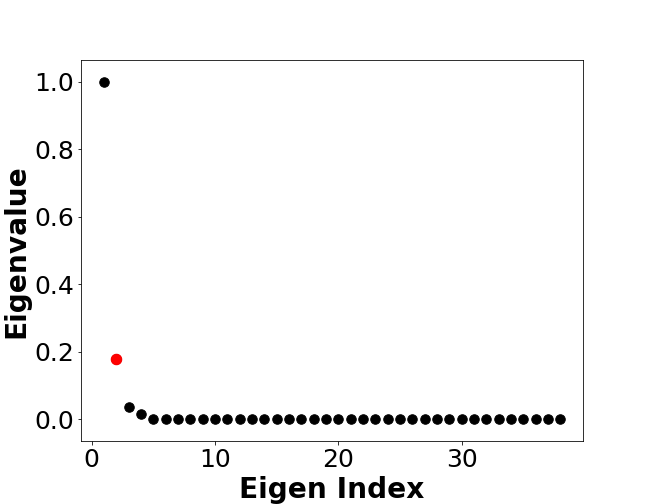}
        \caption{ \footnotesize  Random Sampling}
        \label{fig:heatmapXeonPhiModel}
    \end{subfigure}
        \end{minipage}
    \captionsetup{singlelinecheck = false, justification=justified}
    \caption{  Sorted and normalized eigenvalues of Banknote Authentication dataset with different sampling methods.   }\label{fig:eigenvalues}
    \end{center}

\begin{center}
\begin{minipage}[b]{.55\linewidth}

    \begin{subfigure}[b]{0.49\textwidth}
    \centering
    \captionsetup{justification=centering,skip=-0.2cm}
        \includegraphics[width=\textwidth]{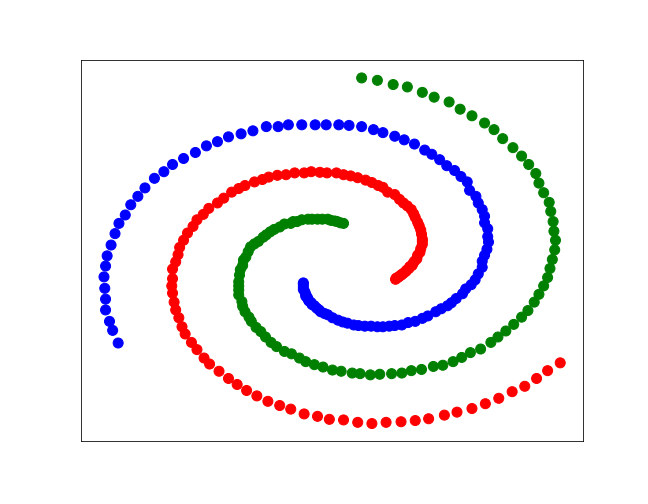}
        \caption{\footnotesize  MM Sampling}
        \label{fig:MM}
    \end{subfigure}
    \begin{subfigure}[b]{0.49\textwidth}
    \centering
    \captionsetup{justification=centering,skip=-0.2cm}
        \includegraphics[width=\textwidth]{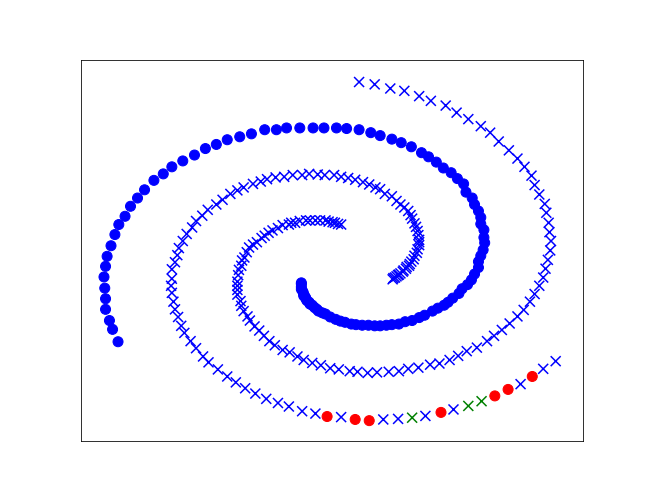}
        \caption{ \footnotesize   k-means Sampling}
        \label{fig:kmeans}
    \end{subfigure}
    \newline
    \begin{subfigure}[b]{0.49\textwidth}
    \centering
    \captionsetup{justification=centering,skip=-0.2cm}
        \includegraphics[width=\textwidth]{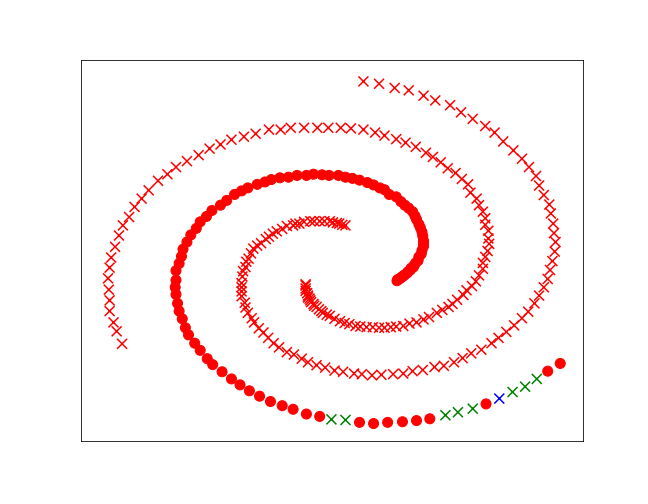}
        \caption{\footnotesize  DPP Sampling}
        \label{fig:DPP}
    \end{subfigure}
    \begin{subfigure}[b]{0.49\textwidth}
    \centering
    \captionsetup{justification=centering,skip=-0.2cm}
        \includegraphics[width=\textwidth]{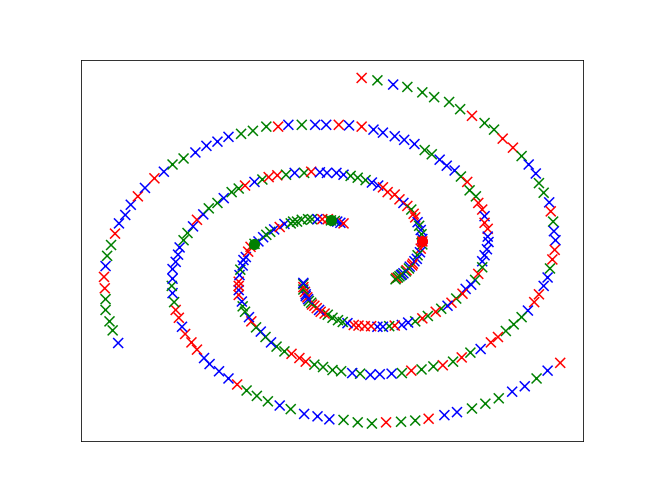}
        \caption{ \footnotesize   Random Sampling}
        \label{fig:rand}
    \end{subfigure}
        \end{minipage}
    \captionsetup{singlelinecheck = false, justification=justified}
    \caption{Qualitative illustration on the \emph{Spiral} dataset by different sampling methods.}\label{fig:spiral}
    \end{center}
\end{figure}

As discussed, to apply GMM, we need embedding of the pairwise \MM \ matrix ($\mathbf{M}$ for original data and $\mathbf{M}_s$ for samples) into a $d^{'}$-dimensional space where $ d^{'} \in \left\{1, \ ..., \texttt{size of dataset} \right\}$. To compute embedding, first, we transform the \MM \ matrix into a Mercer kernel and then perform an eigenvalue decomposition.
Figure \ref{fig:eigenvalues} shows the sorted normalized eigenvalues of the $\mathbf{M}_s$ obtained from the \emph{Banknote Authentication} dataset for different sampling methods. It can be observed that the eigenvalues drop in magnitude after a certain eigen index. This point is shown by red in  Figure \ref{fig:eigenvalues}. We consider this point as the proper value for $d^{'}$; so, the selected $d^{'}$ is adjusted according to the dynamics of eigenvalues (using the elbow rule).

\begin{table}[t]
\begin{center}

\caption{Quantitative results of different sampling schemes.}
 \resizebox{0.95\textwidth}{!}{
 \begin{minipage}
{\textwidth}
\label{tab:result} \centering
\begin{tabular}{c|c|c|ccccc}
 & \multicolumn{1}{|c|}{Dataset} & Metric & None  & MM & k-means & DPP & Random \\

    \hline  \hline
\multirow{9}{*}{\rotatebox{90}{Synthetic Data}} & \multirow{3}{*}{Pathbased} & M1 &62.11\%  & \textbf{ 61.05}\% &  47.27\%& 53.47\% & \hphantom{0}0.22\% \\
 & & M2 & 66.39 \% & \textbf{68.01}\% & 52.20\% & 62.37\% & \hphantom{0}0.10\%\\
 & & M3 & 67.22\% & \textbf{70.20}\% & 55.52\% & 65.81\%  & \hphantom{0}0.74 \% \\
  \cline{2-8}
& \multirow{3}{*}{Spiral}  & M1 & 100\%  & \textbf{100}\% &  \hphantom{0}8.21\%& \hphantom{0}7.57\% & \hphantom{0}0.24\% \\
 & & M2 & 100 \% & \textbf{100}\% &10.55\% & 15.31\% & \hphantom{0}0.26\%\\
 & & M3 & 100\% & \textbf{100}\% & 11.39\% & 17.25\%  & \hphantom{0}0.34 \% \\
  \cline{2-8}
 & \multirow{3}{*}{Aggregation} & M1 &86.87\%  & 80.45\% &  \textbf{92.08}\%& 84.78\% & \hphantom{0}0.93\% \\
 & & M2 & 87.69\% & 80.82\% & \textbf{91.19}\% & 88.81\% & \hphantom{0}1.14\%\\
 & & M3 & 92.31\% & 90.31\% & \textbf{93.26}\% & 92.24\%  & \hphantom{0}2.51\% \\

  \hline \hline
\multirow{15}{*}{\rotatebox{90}{Real Data}} & \multirow{3}{*}{Banknote}& M1 & 58.53\% & \textbf{58.11}\%  &  \hphantom{0}7.01 \% & 12.65\% & \hphantom{0}0.06\% \\
 & & M2 & 58.06\% & \textbf{52.26} \% & \hphantom{0}5.00\% & 16.62\% & \hphantom{0}0.00\% \\
 & & M3 & 58.15\% & \textbf{52.33} \% & \hphantom{0}5.84\% & 21.45\% & \hphantom{0}0.06\% \\
 \cline{2-8}
& \multirow{3}{*}{Cloud} & M1 & 100\% & \textbf{96.32}\% & 73.00\% & 95.28\% & \hphantom{0}0.01\% \\
 & & M2 & 100\% &93.33 \% & 68.53\% & \textbf{ 94.77}\% & \hphantom{0}0.00\% \\
 & & M3 & 100\% & \textbf{93.32}\% & 96.07\% &  93.18 \%& \hphantom{0}0.00\% \\
  \cline{2-8}
 & \multirow{3}{*}{Iris} & M1 & 55.10\% & 66.37\% & \textbf{74.55}\% & 64.10\% & \hphantom{0}0.56\% \\
 & & M2 & 58.26\% & 68.64\% &\textbf{ 78.79}\% & 66.87\% & \hphantom{0}0.40\% \\
 & & M3 & 66.94\% &\textbf{ 71.74}\% & 69.42\% & 70.48\% & \hphantom{0}1.72\% \\
  \cline{2-8}
 & \multirow{3}{*}{Perfume} & M1 & 78.44 \% &\textbf{ 64.88}\% &  59.75\%& 40.04\% & \hphantom{0}0.08\% \\
 & & M2 &89.24\%  & 74.28\% & \textbf{78.78}\% & 51.22\% &  \hphantom{0}0.14\%\\
 & & M3 & 92.48\% & \textbf{85.64}\%  & 81.44\% &  53.88\%& 12.76\% \\
 \cline{2-8}
  & \multirow{3}{*}{Seed}
 & M1 & 61.89\%  &48.26 \% &  \textbf{75.25} \%& 65.52\% & \hphantom{0}0.47\% \\
 & & M2 & 57.55\% & 45.88\% &  \textbf{70.44}\% & 67.08\% & \hphantom{0}0.69\%\\
 & & M3 & 58.29\% & 50.73\% & 69.74\% & \textbf{69.83 }\%  & \hphantom{0}1.73\% \\
  \hline \hline
 \multirow{6}{*}{\rotatebox{90}{Image Data}} & \multirow{3}{*}{ \shortstack{Bottle \\ 2009\_002204 }}
 & M1 & 70.15\%  & \textbf{70.13}\% & 64.39 \%& 53.12\% & \hphantom{0}1.17\% \\
 & & M2 & 56.30\% & \textbf{55.23}\% & 48.05\% & 36.88\% & \hphantom{0}0.37\%\\
 & & M3 & 56.48\% & \textbf{56.33}\% & 52.99\% &41.20\%  & \hphantom{0}0.18\% \\
 \cline{2-8}
  & \multirow{3}{*}{\shortstack{Bird \\ 2011\_001967 } }
 & M1 & 66.89\%  & \textbf{56.33}\% &  54.33\%& 31.88\% & \hphantom{0}1.30\% \\
 & & M2 & 41.62\% & 32.30\% & \textbf{33.50}\% & 18.72\% & \hphantom{0}0.10 \%\\
 & & M3 & 43.23\% & \textbf{37.59}\% & 36.10\%  &20.29\%  & \hphantom{0}0.39\% \\

\end{tabular}
\end{minipage} }
\end{center}
\end{table}

Therefore, we apply GMM to the embedding vectors $\mathbf{E}_{d^{'}}$ and then extend the sample labels to all the other objects represented by the corresponding sample. Finally, as shown in Table \ref{tab:result}, we compare the estimated cluster labels with the ground truth labels. Table \ref{tab:result} shows the quantitative results of the different sampling methods. The \textit{None} column refers to the \emph{no sampling} case, where we use all the objects. In this case, we obtain \MM \ distance matrix for all objects and apply GMM to the respective embedded vectors. Despite memory inefficiency, the results of this case provide information about effectiveness of sampling.

We observe that for two out of three synthetic datasets, MM Sampling outperforms the other methods. However, even for \emph{Aggregation} dataset the results from MM Sampling are acceptable.
Similarly, on UCI datasets, MM Sampling yields often the best or close to best results. Finally, on image segmentation, MM Sampling still achieves the best scores.

Qualitative results on the \emph{Spiral} dataset are illustrated in Figure \ref{fig:spiral}, where different colors denote different estimated clusters, and dots and crosses respectively indicate correctly and incorrectly labeled objects. Figure \ref{fig:MM} shows GMM results obtained by MM Sampling, where all objects are correctly clustered. MM Sampling adapts well with the elongated structures in the data. Figure \ref{fig:kmeans} and Figure \ref{fig:DPP} illustrate the clustering results with $k$-means and DPP sampling methods. Both methods mistakenly assign almost all data to a single cluster. Finally, Figure \ref{fig:rand} shows the results of GMM with random sampling.

In terms of space complexity, random and MM Sampling satisfy a linear space complexity $\bigO \left( N\right)$. $k$-means sampling requires $\bigO \Big( N \big( D$\textit{+}$ K \big) \Big) $ memory, and as discussed, standard DPP requires computing samples offline with the space complexity of $\bigO \left(N^2 \right)$.

\section{Conclusion}

We developed a generic framework for memory-efficient computation of \MM \ distances based on effective sampling schemes.
Within this framework, we developed an adaptive and memory-efficient sampling method consistent with the pairwise \MM \ distances on the entire datasets. We evaluated the framework and the sampling methods on clustering of several datasets with GMM. 

\section*{Acknowledgment}
This work is partially supported by the Wallenberg AI, Autonomous Systems and Software Program (WASP) funded
by the Knut and Alice Wallenberg Foundation.

\bibliographystyle{unsrt}
\bibliography{ref}  

\end{document}